\documentclass{article}

\usepackage[final]{neurips_2024}

\usepackage[utf8]{inputenc} 
\usepackage[T1]{fontenc}    
\usepackage{hyperref}       
\usepackage{url}            
\usepackage{booktabs}       
\usepackage{amsfonts}       
\usepackage{nicefrac}       
\usepackage{microtype}      
\usepackage{xcolor}         
\usepackage{xspace}
\usepackage[pdftex]{graphicx}
\usepackage{subcaption}
\usepackage{makecell}
\usepackage[acronym]{glossaries} 
\robustify{\gls}
\newacronym{DNN}{DNN}{Deep Neural Networks}
\newacronym{DL}{DL}{Deep Learning}
\newacronym{CNN}{CNN}{Convolutional Neural Network}
\newacronym{ML}{ML}{Machine Learning}
\newacronym{NN}{NN}{Neural Network}
\newacronym{ViT}{ViT}{Vision Transformer}
\newacronym{MHA}{MHA}{Multi-head Attention}
\newacronym{MLP}{MLP}{Multilayer Perceptron}
\newacronym{NLP}{NLP}{Natural Language Processing}
\newacronym{NORM}{NORM}{Normalization Layer}
\newacronym{EMA}{EMA}{Exponential Moving Average}
\usepackage{booktabs}
\usepackage{multirow}
\usepackage{graphicx,wrapfig,lipsum}
\usepackage{algorithm2e}
\usepackage{algpseudocode}
\usepackage{mathtools}
\usepackage{nicematrix}
\usepackage{booktabs}
\usepackage{pifont}
\newcommand{\cmark}{\ding{51}}%
\newcommand{\xmark}{\ding{55}}%

\newcommand{\review}[1]{{\color{red}{#1}}}
\renewcommand{\review}[1]{#1}

\newcommand{\name}{\texttt{\textbf{HydraViT}}\xspace}

\title{\name: Stacking Heads for a Scalable ViT}

\author{%
  Janek Haberer\textsuperscript{*}, Ali Hojjat\textsuperscript{*}, Olaf Landsiedel \\
  Kiel University, Germany\\  \textsuperscript{*}\small{\textit{Equal contribution}} \\
  \texttt{\{janek.haberer,ali.hojjat,olaf.landsiedel\}@cs.uni-kiel.de} \\
}

\begin{document}

\maketitle

\begin{abstract}
  The architecture of \glspl{ViT}, particularly the \gls{MHA} mechanism, imposes substantial hardware demands. Deploying \glspl{ViT} on devices with varying constraints, such as mobile phones, requires multiple models of different sizes. However, this approach has limitations, such as training and storing each required model separately. This paper introduces \name, a novel approach that addresses these limitations by stacking attention heads to achieve a scalable \gls{ViT}. By repeatedly changing the size of the embedded dimensions throughout each layer and their corresponding number of attention heads in \gls{MHA} during training, \name induces multiple subnetworks. Thereby, \name achieves adaptability across a wide spectrum of hardware environments while maintaining performance. Our experimental results demonstrate the efficacy of \name in achieving a scalable \gls{ViT} with up to 10 subnetworks, covering a wide range of resource constraints. \name achieves up to 5 p.p. more accuracy with the same GMACs and up to 7 p.p. more accuracy with the same throughput on ImageNet-1K compared to the baselines, making it an effective solution for scenarios where hardware availability is diverse or varies over time. 
  The source code is available at \url{https://github.com/ds-kiel/HydraViT}.
\end{abstract}

\begin{figure}[h]
    \centering
    \begin{subfigure}[b]{0.49\textwidth}
        \centering
        \includegraphics[trim=0 5 0 0, clip, width=0.66\textwidth]{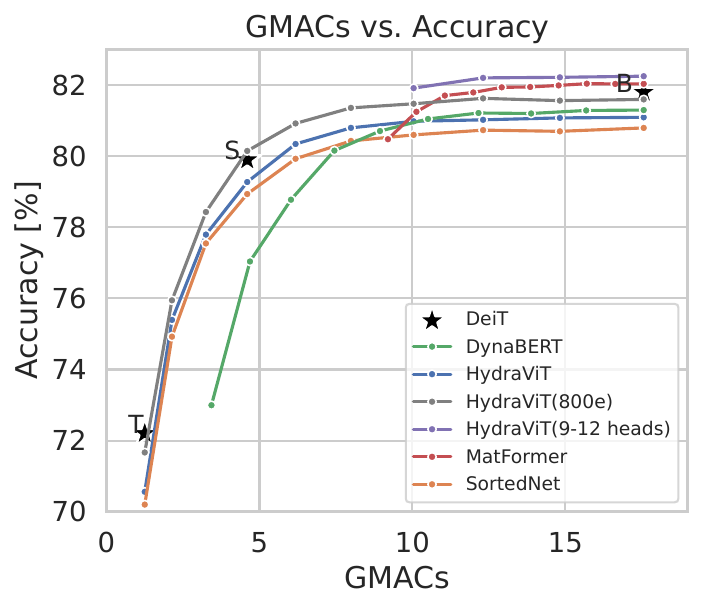}
        \includegraphics[trim=20 5 0 0, clip, width=0.32\textwidth]{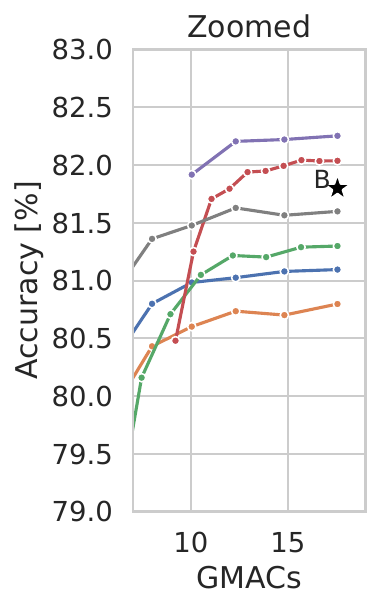}
        \caption{GMACs vs. Accuracy}    
        \label{fig:GMACs_acc_pretrained}
    \end{subfigure}
    \begin{subfigure}[b]{0.5\textwidth}
        \centering
        \includegraphics[trim=0 5 0 0, clip, width=0.65\textwidth]{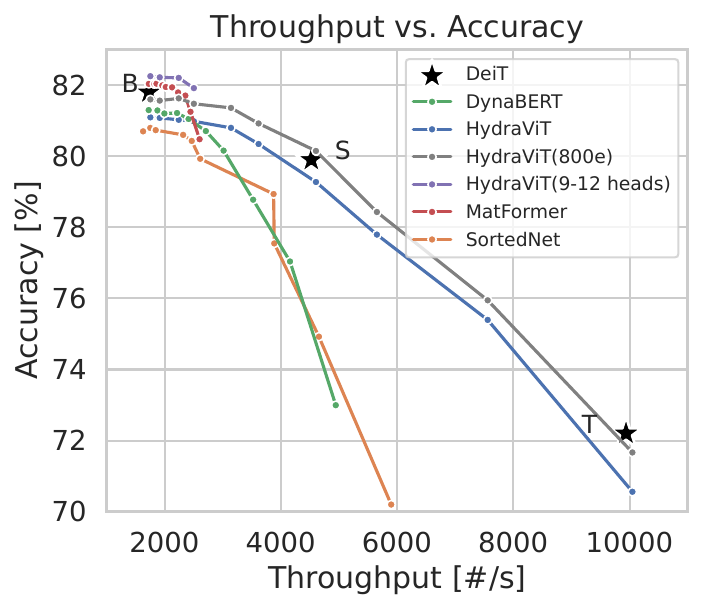}
        \includegraphics[trim=20 0 0 0, clip, width=0.33\textwidth]{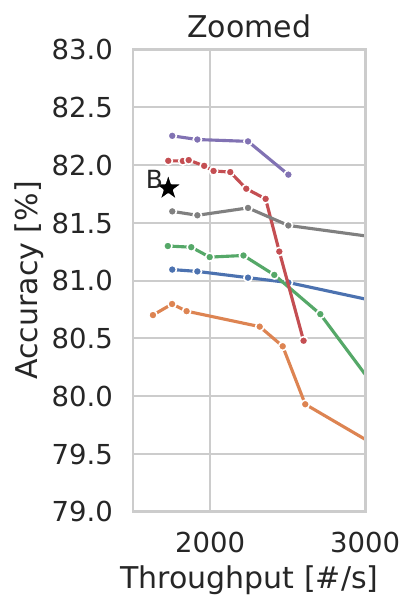}
        \caption{Throughput vs. Accuracy}    
        \label{fig:throughput_acc_pretrained}
    \end{subfigure}
   \caption{Performance comparison of \name and baselines on ImageNet-1K in terms of GMACs (a) and throughput (b) evaluated on NVIDIA A100 80GB PCIe. \name trained on 3-12 heads demonstrates superior performance over DynaBERT \citep{hou2020dynabert} and SortedNet \citep{valipour2023sortednet}. While MatFormer \citep{kudugunta2023matformer} shows higher performance than \name within its limited scalability range, but when we train on a narrower scalability range (9-12 heads), \name surpasses MatFormer. We also show that training \name for more epochs can further improve accuracy. Note that each line corresponds to one model, and changing the number of heads in the vanilla DeiT models significantly drops their accuracy to less than 30\%.}
\label{fig:all_acc_pretrained_GMACS_throughput}
\end{figure}
\newpage 
\section{Introduction}

\paragraph{Motivation} Following the breakthrough of Transformers \citep{vaswani2017attention},
\citet{dosovitskiy2021an} established the \acrfull{ViT} as the base transformer architecture for computer vision tasks. 
As such, numerous studies build on top of \glspl{ViT} as their base \citep{liu2021swin, tolstikhin2021mlp, yu2022metaformer}. 
In this architecture, \acrfull{MHA} plays an important part, capturing global relations between different parts of the input image.
However, \glspl{ViT} have a much higher hardware demand due to the size of the attention matrices in \gls{MHA}, which makes it challenging to find a configuration that fits heterogeneous devices.

\begin{wraptable}{l}{4.5cm}
\tiny
    \centering
    \caption{ViT Configurations}
    \begin{tabular}{lccccc}
        \Xhline{2\arrayrulewidth} \\[-2ex]
         & \textbf{ViT-Ti} & \textbf{ViT-S} & \textbf{ViT-B} \\ 
        \Xhline{2\arrayrulewidth}\\[-2ex]
        \textbf{\# Layers} & 12 & 12 & 12 \\
        \textbf{Dim} & 192 & 384 & 768 \\  
        \textbf{\# Heads} & 3 & 6 & 12 \\
        \textbf{Dim per Head} & 64 & 64 & 64 \\
        \textbf{\# Params} & 5.7 M & 22 M & 86 M \\
        \Xhline{2\arrayrulewidth}\\[-2ex]
    \end{tabular}
    \label{tab:comparison}
\end{wraptable}

To accommodate devices with various constraints, \glspl{ViT} offer multiple independently trained models with different sizes and hardware requirements, such as the number of parameters, FLOPS, MACs, and hardware settings such as latency and RAM, with sizes typically increasing nearly at a logarithmic scale \citep{kudugunta2023matformer}, see Table~\ref{tab:comparison}.
Overall, in the configurations of \glspl{ViT}, the number of heads and their corresponding embedded dimension in \gls{MHA} emerges as the key hyperparameter that distinguishes them.

While being a reasonable solution for hardware adaptability, this approach has two primary disadvantages:
(1) Despite \review{larger models, e.g., ViT-S and ViT-B,} not having a significant accuracy difference, each of these models needs to be individually trained, tuned, and stored, which is not suitable for downstream scenarios where the hardware availability changes over time.
(2) Although the configuration range covers different hardware requirements, the granularity is usually limited to a small selection of models and cannot cover all device constraints.

\paragraph{Observation} By investigating the architecture of these configurations, we notice that ViT-Ti, ViT-S, and ViT-B share the same architecture, except they differ in the size of the embeddings and the corresponding number of attention heads they employ, having 3, 6, and 12 heads, respectively. In essence, this can be expressed as $ViT_{T}\subseteq ViT_{S} \subseteq ViT_{B}$, see Table~\ref{tab:comparison}.

\paragraph{Research question} In this paper, we address the following question: Can we train a universal ViT model with $H$ attention heads and embedding dimension $E$, such that by increasing the embedded dimension from $e_{1}$ to $e_{2}$, where $e_{1} < e_{2} \leq E$, and its corresponding number of heads from $h_{1}$ to $h_{2}$, where $h_{1} < h_{2} \leq H$, the model's accuracy gracefully improves?

\paragraph{Approach}

\begin{wrapfigure}{r}{3cm}
\includegraphics[trim=5 5 10 25, width=3cm]{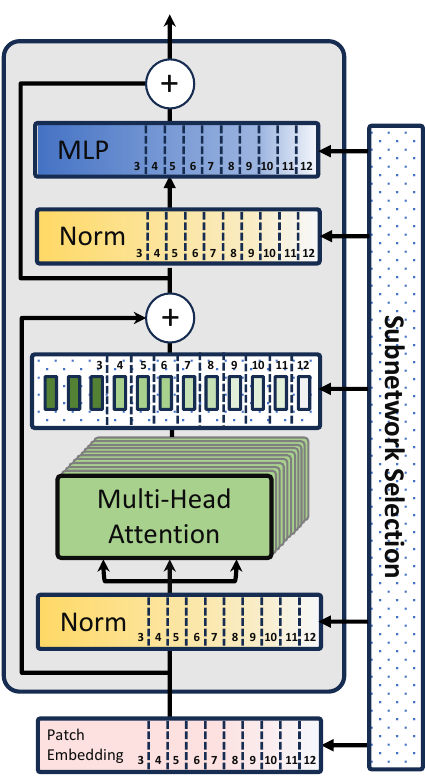}
\caption{Architecture of \name}
\vspace{-1em}
\label{fig:HydraViT}
\end{wrapfigure}
In this paper, we propose \name, a stochastic training approach that extracts subsets of embeddings and their corresponding heads within \gls{MHA} across a universal \gls{ViT} architecture and jointly trains them.
Specifically, during training, we utilize a uniform distribution to pick a value $k$, where $k \leq H$. Subsequently, we extract the embedded dimension ($[0:k \times HeadDim]$), where $HeadDim$ is the size of each head, and its corresponding first $k$ heads ($[0:k]$) and only include these in both the backpropagation and forward paths of the training process.
To enable the extraction of such subnetworks, we reimplement all components of the \gls{ViT} including \gls{MHA}, \gls{MLP}, and \gls{NORM}, see Fig.~\ref{fig:HydraViT}.
By using this stochastic approach, the heads will be stacked based on their importance, such that the first heads capture the most significant features and the last heads the least significant ones from the input image.

After the training phase is completed, during inference, \name can dynamically select the number of heads based on the hardware demands. For example, if only $p\%$ of the hardware is available, \name extracts a subnetwork with the embedded size of $\lceil p \times H \rceil\times HeadDim $ and the first $\lceil p \times H \rceil$ heads and runs the inference.
This flexibility is particularly advantageous in scenarios such as processing a sequence of input images, like a video stream, where latency is critical, especially on constrained devices such as mobile devices. In such environments, where various tasks are running simultaneously, and hardware availability dynamically fluctuates, or we need to meet a deadline, the ability to adapt the model's configuration without loading a new model offers significant benefits.

\textbf{Contributions:}
\begin{enumerate}
\itemsep0em

\item We introduce \name, a stochastic training method that extracts and jointly trains subnetworks inside the standard \gls{ViT} architecture for scalable inference.
\item In a standard \gls{ViT} architecture with $H$ attention heads, \name can induce $H$ submodels within a universal model.
\item \name outperforms its scalable baselines with up to 7 p.p. more accuracy at the same throughput and performance comparable to the respective standard models DeiT-tiny, DeiT-small, and DeiT-base, see Figure~\ref{fig:all_acc_pretrained_GMACS_throughput} for details.

\end{enumerate}

\begin{figure*}
    \centering
    \begin{subfigure}[b]{1.\textwidth}
        \centering
        \includegraphics[trim=5 0 0 0, clip, width=1\textwidth]{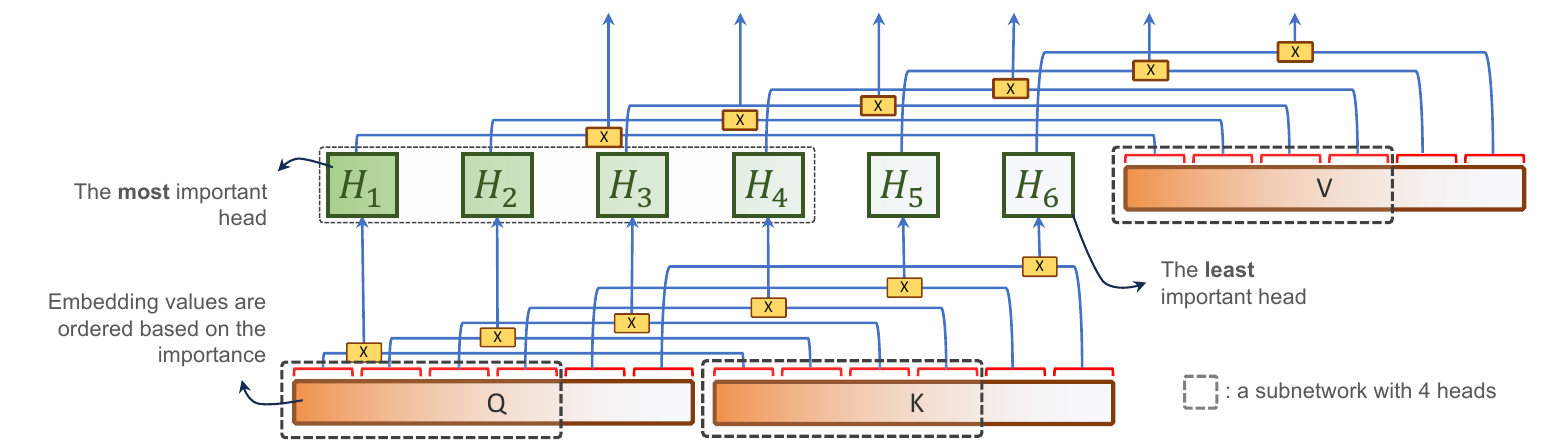}
    \end{subfigure}
    \caption{In this figure, we illustrate an example of how we extract a subnetwork with 4 heads in \gls{MHA} with a total number of 6 heads.
    In \name, with the stochastic dropout training, we order the attention heads in \gls{MHA} and consequently their corresponding embedding vectors based on their importance.}  
\label{fig:Attention}
\end{figure*}

\section{Related Work}
\label{sec:rw}
The original \acrfull{ViT}~\citep{dosovitskiy2021an} has become the default architecture of Transformer-based vision models. While many works improve upon the original implementation by changing the architecture or training process~\citep{liu2022swin, touvron2022deit, wang2021pyramid}, none of these works yield a scalable architecture and need multiple separate sets of weights to be able to deploy an efficient model on devices with various constraints.

For \glspl{CNN}, \citet{fang2018nestdnn} create a scalable network by pruning unimportant filters and then repeatedly freezing the entire model, adding new filters, and fine-tuning. Thereby, they achieve a network that can be run with a flexible number of filters. \citet{yu2018slimmable} achieve the same, but instead of freezing, they train a network for different layer widths at once.
For Transformers, \citet{chavan2022vision} use sparsity to efficiently search for a subnetwork but then require fine-tuning for every extracted subnetwork to acquire good accuracy.

\citet{beyer2023flexivit} introduce a small change in the training process by feeding differently sized patches to the network. Thereby, they can reduce or increase the number of patches, affecting the speed and accuracy during inference. Other works use the importance of each patch to prune the least important patches during inference to achieve a dynamic \gls{ViT} \citep{yin2022vit, rao2021dynamicvit, tang2022patch, wang2021not}.

Matryoshka Representation Learning~\citep{kusupati2022matryoshka} and Ordered Representations with Nested Dropout~\review{\citep{rippel2014learning, Hojjat_2023_CVPR}} are techniques to make the embedding dimension of Transformers flexible, i.e., create a Transformer, which can also run partially. \citet{kudugunta2023matformer} use Matryoshka Learning to make the hidden layer of the MLP in each Transformer block flexible. \citet{hou2020dynabert} change the hidden layer of the \gls{MLP} and the \gls{MHA} but still use the original dimension between Transformer blocks and also between \gls{MHA} and \gls{MLP}. \citet{salehi2023sharcs} make the entire embedding dimension in a Transformer block flexible. However, they rely on a few non-flexible blocks followed by a router that determines the embedding dimension for the flexible blocks, which adds complexity and hinders the ability to choose with which network width to run.

\citet{valipour2023sortednet} propose SortedNet that trains networks to be flexible in depth and width. However, they mainly focus on evaluating with \glspl{CNN} on CIFAR10~\citep{krizhevsky2009learning} and Transformers on \gls{NLP} tasks in contrast to us. Additionally, they keep the number of heads in \gls{MHA} fixed at 12, \review{whereas we show that reducing the number of heads coupled to the embedding dimension, i.e., the first 64 values of the embedding always belong to the first head in \gls{MHA}, the second 64 values belong to the second head, and so on, removes inconsistencies in the scaling of the \gls{MHA} and improves performance.}

Motivated by these previous works, in \name, we propose a flexible \gls{ViT} in which we, unlike previous works, adjust every single layer, and except for one initial training run, there is no further fine-tuning required. Additionally, we show that reducing the number of heads coupled to the embedding dimension, a weighted subnetwork sampling distribution, and adding separate classifier heads improves the performance of subnetworks.

\section{\name}\label{sec:design}

In this section, we introduce \name, which builds upon the \gls{ViT} architecture. We start by detailing how general \glspl{ViT} function. Next, we explain how \name can extract subnetworks within the \gls{MHA}, \gls{NORM}, and \gls{MLP} layers. Finally, we describe the stochastic training regime used in \name to simultaneously train a universal \gls{ViT} architecture and all of its subnetworks.

\begin{figure*}
    \centering
    \begin{subfigure}[b]{0.69\textwidth}
        \centering
        \includegraphics[trim=0 0 0 0, clip, width=1\textwidth]{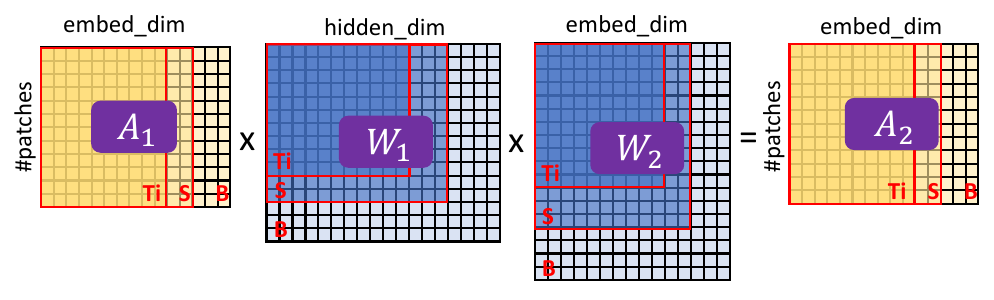}
        \caption{\gls{MLP}}    
        \label{fig:MULT}
    \end{subfigure}
    \begin{subfigure}[b]{0.30\textwidth}
        \centering
        \includegraphics[width=1\textwidth]{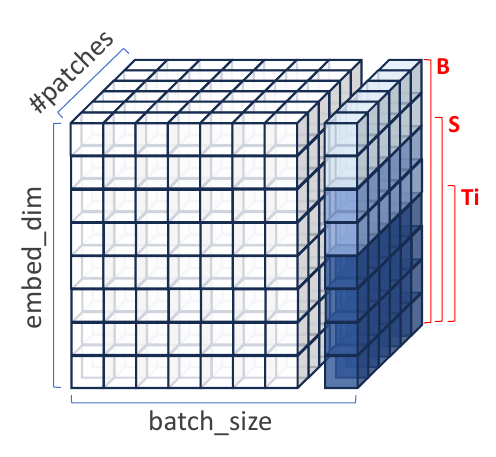}
        \caption{\gls{NORM}}     
        \label{fig:NORM}
    \end{subfigure}
   \caption{An illustration of subnetwork extraction within \gls{MLP} and \gls{NORM} layers, introduced in \name.
   Fig.~\ref{fig:MULT} demonstrates how \name slices activations, denoted as $A_{1}$ and $A_{2}$, along with their respective weight matrices, denoted as $W_{1}$ and $W_{2}$, based on the number of utilized heads.
   Also, Fig.~\ref{fig:NORM} shows how \name applies normalization on the activation corresponding to the used heads.
   For simplicity, only subnetworks with 3, 6, and 12 heads, corresponding to ViT-Ti, ViT-S, and ViT-B respectively, are presented.}
\label{fig:LAYERS}
\end{figure*}

\paragraph{Vision Transformer}
\name is based on the \gls{ViT} architecture \citep{dosovitskiy2021an}.
We start by taking the input image $x$ and breaking it down into $P$ patches.
Each patch is then embedded into a vector of size $E$ using patch embedding, denoted as $\mathcal{E}^{E}$. Positional encoding is subsequently applied to the embeddings.
Following these preprocessing steps, it passes the embeddings through $L$ blocks consisting of \gls{MHA} with $H$ heads denoted as $\mathcal{A}^{H}$,  \gls{NORM} layer denoted as $\mathcal{N}^{P}$, \gls{MLP} denoted as $\mathcal{M}^{E \times M \times E}$ to predict the class of the input image $x$, where $M$ is the dimension of the hidden layer of the \gls{MLP}. With the model parameters $\theta$, we can formulate this architecture as follows:
\begin{equation}
V_{\theta}(x;\mathcal{E}^{E};\mathcal{A}^{H};\mathcal{M}^{E \times M \times E};\mathcal{N}^{P})
\label{eq:0}
\end{equation}
\paragraph{HydraViT}
\name is able to induce any subnetwork with $k \le H$ heads within the standard architecture of \gls{ViT}.
To do so, \name extracts the first $k$ heads denoted as $\mathcal{A}^{[0:k]}$, and the embeddings corresponding to these heads in \gls{MHA} and \gls{NORM} layers.
Additionally, it extracts the initial $[\frac{E}{H} \times k]$ neurons from the first and last layers of the \gls{MLP}, and the first $[\frac{M}{H} \times k]$ neurons from the hidden layer of \gls{MLP}. Therefore, we can formulate the subnetwork extracted from Eq.~\ref{eq:0} as follows:
\begin{equation}
  V_{\theta_{k}}(x;
  \mathcal{E}^{[0: (\frac{E}{H} \times k) ]};
  \mathcal{A}^{[0:k]};
  \mathcal{M}^{[0: (\frac{E}{H} \times k)]  \times [0: (\frac{M}{H} \times k)]  \times [0: (\frac{E}{H} \times k)]};
  \mathcal{N}^{[0: (\frac{E}{H} \times k)]}); k \in \{1,2, \dots, H\}
\label{eq:1}
\end{equation}

\begin{figure}
\begin{minipage}{.65\textwidth}
\RestyleAlgo{ruled}
\begin{algorithm}[H]
    \caption{Stochastic dropout training}\label{alg:alg1}
    \KwData{\name: $V_{{\theta}_{k}}$,\\
    Number of batches: $N_{batch}$,\\
    Number of the heads of the universal model: $H$,\\
    Uniform distribution:  $\mathcal{U}$.}
    \For {$ 1 \leq e_{i} \leq N_{epoch}$}{
      \For {$ 1 \leq b_{i} \leq N_{batch}$}{
      \vspace{0.2em}
            \textcolor{lightgray}{\textbackslash* sample a subnetwork *\textbackslash} \\
      \vspace{0.2em}
            $V_{\theta} \xrightarrow[]{k \sim  \mathcal{U}(k)}V_{{\theta}_{k}}, k \in \{1,2, \dots H\}$; \\
      \vspace{0.2em}
            \textcolor{lightgray}{\textbackslash* calculate single-objective loss *\textbackslash} \\
      \vspace{0.2em}
            $\mathcal{L}(V_{{\theta}_{k}}(x_{b_{i}}), y)$; \\
      \vspace{0.2em}
            Back-propagation through subnetwork $V_{{\theta}_{k}}$; \\
       }
    }
\end{algorithm}
\end{minipage}
\hfill
\begin{minipage}{.3\textwidth}
\includegraphics[trim=0 0 0 0, width=4cm]{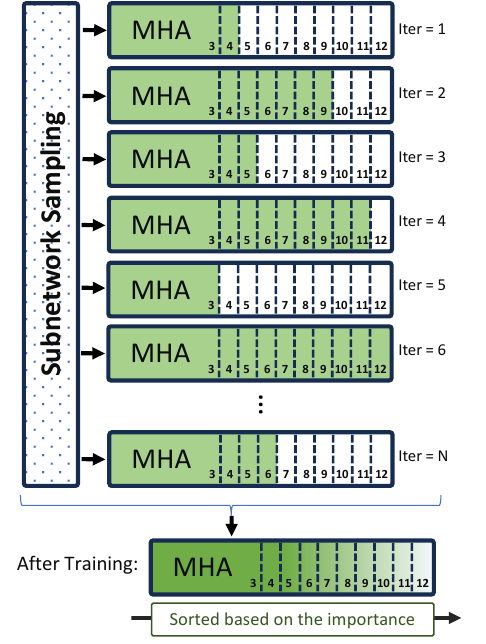}
\caption{Stochastic tail-drop training.}
   \label{fig:training}
\end{minipage}
\end{figure}

Figure~\ref{fig:LAYERS} illustrates how \name extracts subnetworks within \gls{NORM} and \gls{MLP} layers. For simplicity, we demonstrate subnetworks with 3, 6, and 12 heads, representing configurations for ViT-Ti, ViT-S, and ViT-B, respectively. Additionally, in Figure \ref{fig:Attention}, we present an example of how \name extracts a subset of heads and their corresponding embeddings in \gls{MHA} layers. \review{By designing \name this way, we can deploy only a subnetwork, e.g., \name with 6 heads, and still have the option at runtime to run with even fewer heads. It is not necessary to deploy \name with all weights, which is necessary for deployment on more constrained devices.}

\paragraph{Stochastic dropout training}
Ideally, to achieve a truly scalable model, we need to extract all the possible subnetworks, calculate their loss, sum them up, and minimize it. This yields the following multi-objective optimization problem:
\begin{equation}
  \min _{[{\theta}_{1} \dots {\theta}_{H}]} \sum_{i=1}^{N}  \sum_{h=1}^H \mathcal{L}(V_{{\theta}_{h}}(x_{i}), y_{i})
\label{eq:2}
\end{equation}

where N is the number of samples, $x_{i}$ is the input and $y_{i}$ is the ground truth. However, optimizing this multi-objective loss function has a complexity of $\mathcal{O}(N \times H)$ and requires at least $H$ times more RAM compared to a single-objective loss function to store the gradient graphs, a demand that exceeds the capacity of a current GPU. To address this issue, we suggest employing stochastic training: On each iteration, instead of extracting all of the $H$ possible subnetworks and optimizing a multi-objective loss function, we sample a value $k \in \{1,2, \dots, H\}$ based on a uniform discrete probability distribution $\mathcal{U}(k)$. Then we extract its respective subnetwork $V_{{\theta}_{k}}$, and minimize only this loss function, see Alg.~\ref{alg:alg1}. 
This approach decreases the complexity of Eq.~\ref{eq:2} to $\mathcal{O}(N)$.
In this training regime, the first parts of embeddings and their corresponding attention heads become more involved in the training process, while the later parts are less engaged. After training, due to this asymmetric training, \review{which can also be seen as an order-aware biased version of dropout,} the embedding values and their respective attention heads are ordered based on importance, see Fig.~\ref{fig:training}. \review{Note that despite the similarity to dropout we do not need scaling during training as our training and inference phases are identical.}
Thereby, we can simplify the Eq.~\ref{eq:2} as follows:

\begin{equation}
  k \sim  \mathcal{U}(k);\quad \min _{{\theta}_{k}} \sum_{i=1}^N \mathcal{L}(V_{{\theta}_{k}}(x_{i}), y_{i})
  \label{eq:important}
\end{equation}

\paragraph{Separate classifiers}
We implement a mechanism to train separate classifier heads for each subnetwork. This adds a few parameters to the model, but only during training or when running the model in a dynamic mode, i.e., having the ability to freely choose for each input with how many heads to run the model. The advantage is that we do not need to find a shared classifier that can deal with the different amounts of features each subnetwork provides. However, if we fix the number of epochs, each classifier gets fewer gradient updates than the shared one, which is why we only use this when training \name with 3 subnetworks.

\paragraph{Subnetwork sampling function}
When trying to train a single set of weights containing multiple subnetworks, we expect an accuracy drop compared to if each subnetwork had its own set of weights. While we mention that we use a uniform discrete probability distribution to sample subnetworks, we can also use a weighted distribution function. With weighted subnetwork sampling, we can guide the model to focus on certain submodels more than others. This is useful in a deployment scenario in which we have many devices with similar resources and want to maximize accuracy for them while maintaining good accuracy for other devices with different resources.

\section{Evaluation}
\label{sec:eval}

\begin{table}
\small
    \caption{The accuracy of \name with our different design choices. "3 Heads" corresponds to a subnetwork that has the same architecture as DeiT-tiny, "6 Heads" corresponds to DeiT-small, and "12 Heads" corresponds to DeiT-base.}
    \centering
    \vspace{0.4em}
    \begin{tabular}{cccrrr}
        \Xhline{2\arrayrulewidth} \\[-2ex]
        \textbf{Weighted} & \textbf{Separate} & \multirow{2}{*}{\textbf{Epochs}} & \textbf{Acc [\%]} & \textbf{Acc [\%]} & \textbf{Acc [\%]} \\ 
        \textbf{Sampling?}& \textbf{Classifiers?} & & \textbf{3 Heads} & \textbf{6 Heads} & \textbf{12 Heads} \\ 
        \Xhline{2\arrayrulewidth}\\[-2ex]
        
        \xmark & \xmark & 300 & 72.56 & 79.35 & 80.63 \\
        \xmark & \xmark & 400 & 73.16 & 79.63 & 80.90 \\
        \xmark & \xmark & 500 & 73.54 & 80.09 & 81.30 \\
        \hline  & \\[-2ex]
        
        \cmark & \xmark & 300 & 72.02 & 79.35 & 80.98 \\
        \cmark & \xmark & 400 & 72.45 & 79.85 & 81.49 \\
        \cmark & \xmark & 500 & 72.50 & 79.89 & 81.63 \\
        \hline  & \\[-2ex]

        \xmark & \cmark & 300 & 72.78 & 79.44 & 80.52 \\
        \xmark & \cmark & 400 & 73.24 & 79.88 & 81.13 \\
        \xmark & \cmark & 500 & 73.42 & 80.12 &  81.13 \\
        \hline  & \\[-2ex]

        \cmark & \cmark & 300 & 72.13 & 79.45 & 81.18 \\
        \cmark & \cmark & 400 & 72.46 & 79.93 & 81.58 \\
        \cmark & \cmark & 500 & 72.65 & 80.08 & 81.77 \\
        \hline  & \\[-2ex]
        
        \multicolumn{3}{c}{DeiT-tiny/small/base} & 72.2 & 79.9 & 81.8 \\

        \Xhline{2\arrayrulewidth}  & \\[-2ex]
    \end{tabular}
    \label{tab:ablation}
\end{table}

In this section, we evaluate the performance of \name and compare it to the baselines introduced in Sec.~\ref{sec:rw}. We assess all experiments and baselines on ImageNet-1K~\citep{5206848} at a resolution of $224 \times 224$. We implement on top of timm~\citep{rw2019timm} and train according to the procedure of \citet{touvron2021training} but without knowledge distillation. 
We use an NVIDIA A100 80GB PCIe to measure throughput. For RAM, we measure the model and forward pass usage with a batch size of 1. We also calculate GMACs with a batch size of 1, i.e., the GMACs needed to classify a single image. 

For the experiments, we used an internal GPU cluster, and each epoch took around 15 minutes. During prototyping, we estimate that we performed an additional 50 runs with 300 epochs.

First, we show that we can attain one set of weights that achieves very similar results as the three separate DeiT models DeiT-tiny, DeiT-small, and DeiT-base~\citep{touvron2021training}. Then, we look at how our design choices, i.e., changing the number of heads coupled to the embedding dimension, weighted subnetwork sampling, and adding separate classifiers for each subnetwork, impact the accuracy.
Afterward, we compare \name to the following three baselines:
\begin{itemize}
    \item \textbf{MatFormer} \citet{kudugunta2023matformer} focus only on the hidden layer of the \gls{MLP} to achieve a flexible Transformer and do not change the heads in \gls{MHA} or the dimension of intermediate embeddings.
    \item \textbf{DynaBERT} \citet{hou2020dynabert} adjust the heads in \gls{MHA} in addition to the dimension of \gls{MLP} and, as a result, make both flexible. However, the intermediate embedding dimension is the same as the original one in between each Transformer block and between \gls{MHA} and \gls{MLP}, which results in more parameters and MACs.
    \item \textbf{SortedNet} \citet{valipour2023sortednet} change every single embedding, including the ones between \gls{MHA} and \gls{MLP} and between Transformer blocks. However, they keep the number of heads in \gls{MHA} fixed, resulting in less information per head \review{and introducing inconsistencies in the scaling of the heads in \gls{MHA}}.
\end{itemize}
In contrast, instead of keeping the number of heads fixed, we change it coupled to the embedding dimension, such that each head gets the same amount of information as in the original \gls{ViT}. We also evaluate adding separate classifiers and employing weighted subnetwork sampling during the training.
Finally, we perform an attention analysis on our model to showcase the effect of adding heads in \gls{MHA}.

\subsection{One set of weights is as good as three: Tiny, Small, and Base at once}
\label{sec:eval_tsb}

\begin{figure}
\begin{minipage}{.45\textwidth}
\centering
    \includegraphics[width=0.9\textwidth]{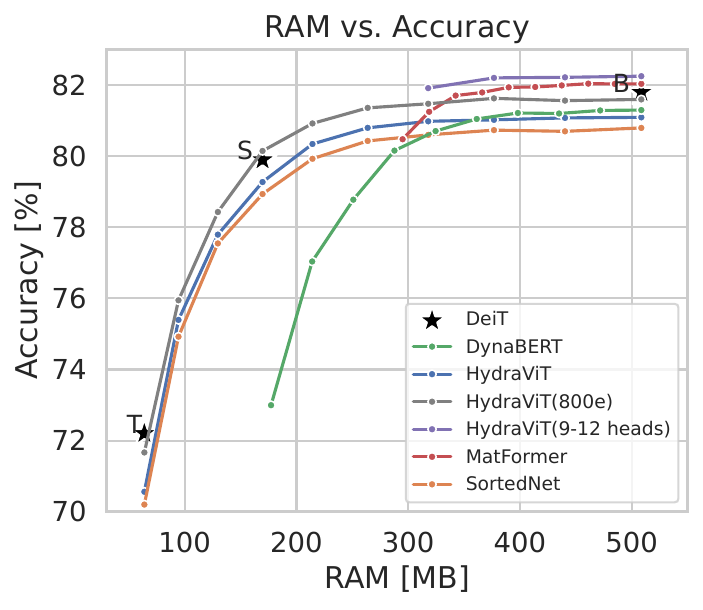}
    \caption{Performance comparison of \name and baselines in terms of RAM usage. Similar to Figure \ref{fig:all_acc_pretrained_GMACS_throughput}, \name achieves the best accuracy per RAM usage. \label{fig:RAM_acc_pretrained}}
\end{minipage}
\hfill
\begin{minipage}{.45\textwidth}
\centering
\vspace{.9em}
    \includegraphics[trim=5 5 5 5, clip, width=1.1\textwidth]{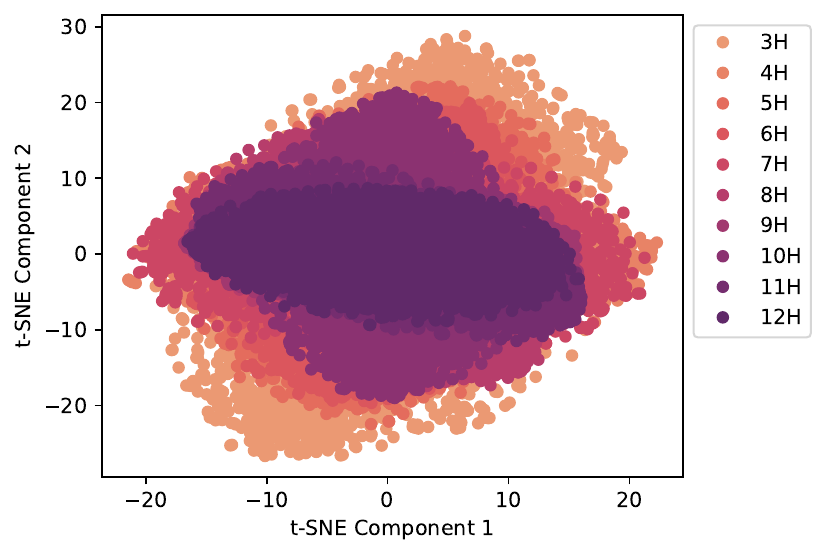}
    \caption{t-SNE representation of the last layer of \name with different numbers of heads. As we can see, having more heads leads to more compact representations.}
\label{fig:tsne}
\end{minipage}
\end{figure}

For this experiment, we train \name for 300, 400, and 500 epochs with a pre-trained DeiT-tiny checkpoint. We show how our design choices, i.e., changing the number of heads coupled to the embedding dimension, weighted subnetwork sampling, and adding separate heads for each subnetwork, impact accuracy. Table \ref{tab:ablation} shows each subnetwork's accuracy for all the combinations of our design choices. Note that subnetworks of \name with 3 heads result in the same architecture as DeiT-tiny, subnetworks with 6 heads result in the same as DeiT-small, and subnetworks with 12 heads result in the same as DeiT-base. 

To evaluate weighted subnetwork sampling, we show in Table~\ref{tab:ablation} that with 25\% weight for training the subnetwork with 3 heads, 30\% weight for 6 heads, and 45\% weight for 12 heads, we can achieve an improvement of 0.3 to nearly 0.6 p.p. for the subnetwork with 12 heads depending on the number of epochs compared to uniform subnetwork sampling. Meanwhile, we get a change of -0.2 to +0.2 p.p. for the subnetwork with 6 heads and a decrease of 0.5 to 1.0 p.p. for the subnetwork with 3 heads compared to uniform subnetwork sampling. Therefore, we can increase accuracy at 12 heads at the cost of an overall accuracy decrease. Keep in mind that removing only one head in the vanilla DeiT-base significantly drops its accuracy
to less than 30\%, whereas \name achieves more than 72\% at 3 heads and 79\% at 6 heads and is therefore more versatile.

To evaluate separate classifiers for each subnetwork, we show in Table~\ref{tab:ablation} that it helps, in some cases, improve each subnetwork's accuracy by up to 0.2 percentage points. But it can also reduce the overall accuracy because each classifier gets fewer gradient updates than a shared classifier.

Finally, we can combine weighted subnetwork sampling and separate classifiers to achieve a high 12-head accuracy, reaching up to 81.77\% accuracy at 500 epochs while maintaining a good accuracy at 3 and 6 heads. We notice that compared to only weighted subnetwork sampling, all the accuracies are up to 0.15 p.p. higher. Due to starting with a pre-trained DeiT-tiny, the classifier for 3 heads needs fewer gradient updates, and the weighted subnetwork sampling shifts the gradient bias to the larger subnetworks, which leads to overall better accuracy, see Table~\ref{tab:ablation}.

To summarize, we show that with \name, we can create one set of weights that achieves, on average, the same accuracy as the three separate models DeiT-tiny, Deit-small, and DeiT-base. To attain this one set of weights, we need at least 300 fewer training epochs than are necessary to train the three different DeiT models. The subnetworks have identical RAM usage, throughput, MACs, and model parameters compared to the DeiT models. While in this section, we investigated \name with only 3 subnetworks, we evaluate \name with more subnetworks in the next section.

\subsection{\name vs. Baselines} \label{sec:Baselines}

For the next experiment, we train \name and the baselines introduced at the beginning of this section for 300 epochs, once from scratch and once with DeiT-tiny as an initial checkpoint. While all of these baselines reduce the embedding dimension, the difference is they reduce it in different parts of the model. We choose 10 subnetworks for each model, setting the embedding dimension from 192 to 768 with steps of 64 in between. These steps correspond to having from 3 to 12 attention heads, with steps of 1 in between. While \name supports up to 12 subnetworks, we choose to exclude the two smallest ones, as their accuracy drops too much.

\begin{table}
  \caption{Comparison of \name with the baselines MatFormer, DynaBERT, and SortedNet. The table shows for selected subnetworks the RAM usage, MACs, model parameters, throughput and their corresponding accuracy when trained from scratch (when applicable) and from the initial DeiT-tiny checkpoint. Note that DynaBERT relies on Knowledge Distillation in every block, which is why it reaches less than 1\% accuracy when trained from scratch.}
  \label{tab:baselines}
  \vspace{0.4em}
  \centering
  \tiny
  \begin{tabular}{lrrrrrrr}
    \toprule
    \multirow{2}{*}{\textbf{Method}}&\multirow{2}{*}{\textbf{Dim}}&\textbf{RAM}&\textbf{MACs}&\textbf{Params}&\textbf{Throughput}&\textbf{Acc [\%]}&\textbf{Acc [\%]}\\
    &&\textbf{[MB]}&\textbf{[G]}&\textbf{[M]}&\textbf{[\# / s]}&\textbf{from scratch}&\textbf{from DeiT-tiny}\\
    \midrule
    \multirow{2}{*}{MatFormer}                      & 768 & 508.45 & 17.56 & 86.6 & 1728.3 & 81.89 & 82.04 \\ 
    \multirow{2}{*}{\citep{kudugunta2023matformer}} & 384 & 366.08 & 11.99 & 58.2 & 2231.6 & 81.52  & 81.80 \\ 
                                                    & 192 & 294.9  & 9.2   & 44.1 & 2601.4 & 79.40 & 80.48  \\
    \midrule
    \multirow{2}{*}{DynaBERT}                       & 768 & 508.45 & 17.56 & 86.6 & 1725.7 & - & 81.30   \\ 
    \multirow{2}{*}{\citep{hou2020dynabert}}        & 384 & 287.62 & 7.45  & 44.1 & 3014.6 & - & 80.16  \\ 
                                                    & 192 & 177.2  & 3.43  & 22.8 & 4944.5 & - & 73.00 \\
    \midrule
    \multirow{2}{*}{SortedNet}                      & 768 & 508.45 & 17.56 & 86.6 & 1753.0 & 79.71 & 80.80 \\ 
    \multirow{2}{*}{\citep{valipour2023sortednet}}  & 384 & 169.6  & 4.6   & 22.1 & 3874.8 & 77.79 & 78.94 \\ 
                                                    & 192 & 63.87  & 1.25  & 5.7  & 5898.2 & 66.64 & 70.20   \\
    \midrule
    \multirow{3}{*}{\name}                          & 768 & 508.45 & 17.56 & 86.6 & 1754.1  & 80.45 & 81.10 \\ 
                                                    & 384 & 169.6  & 4.6   & 22.1 & 4603.6  & 78.40 & 79.28 \\ 
                                                    & 192 & 63.87  & 1.25  & 5.7  & 10047.6 & 67.34 & 70.56  \\
    \midrule
    \multirow{2}{*}{\name}                          & 768 & 508.45 & 17.56 & 86.6 & 1754.1  & \review{81.93} & 81.60   \\ 
    \multirow{2}{*}{800 Epochs}                     & 384 & 169.6  & 4.6   & 22.1 & 4603.6  & \review{79.84} & 80.15  \\ 
                                                    & 192 & 63.87  & 1.25  & 5.7  & 10047.6 & \review{68.78} & 71.67 \\
    \midrule
    \multirow{3}{*}{\name}                          & 768 & 508.45 & 17.56 & 86.6 & 1754.1 & 81.56 & 82.25 \\ 
    \multirow{3}{*}{9-12 Heads}                     & 704 & 440.19 & 14.82 & 72.9 & 1916.1 & 81.55 & 82.22 \\ 
                                                    & 640 & 376.63 & 12.31 & 60.3 & 2242.9 & 81.51 & 82.21 \\
                                                    & 576 & 317.8  & 10.04 & 49.0 & 2385.2 & 81.21 & 81.92 \\
    \bottomrule
  \end{tabular}
\end{table}

Table~\ref{tab:baselines} shows how each baseline compares to \name relative to their RAM usage, GMACs, model parameters, and throughput when training from scratch and when starting with a pre-trained DeiT-tiny checkpoint. Figure~\ref{fig:all_acc_pretrained_GMACS_throughput} and Figure~\ref{fig:RAM_acc_pretrained} show the results of all subnetworks when starting with a pre-trained DeiT-tiny checkpoint. Besides \name, only SortedNet can run with less than 150 MB of RAM while achieving on average 0.3 to 0.7 p.p. worse accuracy than \name. The other baselines, which have a more limited range of subnetworks, achieve a better accuracy when running at higher embedding dimensions. The limited range, however, has the downside of not having smaller subnetworks for devices with fewer resources. And if we limit \name to a similar range as MatFormer, training on 9 to 12 heads, we show that \name reaches the overall highest accuracy at 82.25\% compared to MatFormer's 82.04\%. We also notice that \name cannot reach the exact same performance as the three DeiT models. This is because training for 10 subnetworks with a shared classifier for only 300 epochs has its toll on the overall performance. One option is to train longer, which we demonstrated for \name with 3 subnetworks in Section \ref{sec:eval_tsb}. We repeat the same here and train \name for 800 epochs, showing that even with 10 subnetworks, we can still achieve near-similar performance as the three different DeiT models. This is while having another 7 subnetworks with similar accuracy per resource trade-off points in between. \review{One caveat, however, is that when training from scratch, \name struggles to get a good accuracy at 3 heads. This is most likely due to a sampling bias as the subnetworks with one and two heads are not included in training and due to training hyperparameters as they differ when training DeiT-tiny compared to DeiT-base.} For detailed results, e.g., each subnetwork for every baseline, See Table~\ref{tab:baselines_appendix} in Appendix~\ref{sec:appendix_baselines}.

In summary, \name achieves, on average, better accuracy than its baselines except for MatFormer within its limited
scalability range. However, we show that training \name on a similar scalability range outperforms MatFormer.

\subsection{Analyzing \name's inner workings}

\begin{figure*}[t!]
  \centering
      \centering
    \begin{subfigure}[b]{1\textwidth}
        \centering
        \includegraphics[trim=0 0 0 0, clip, width=1\textwidth]{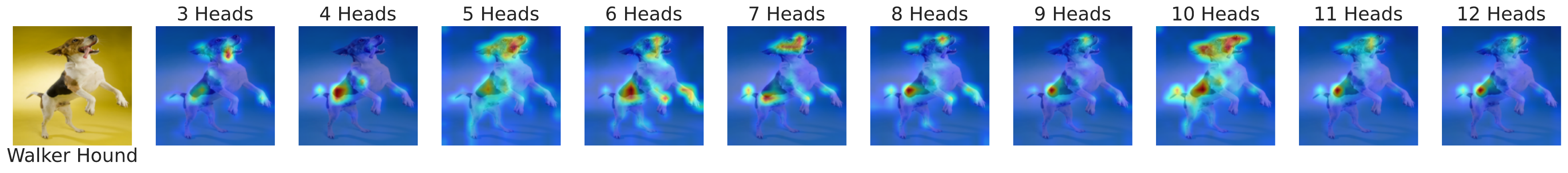}
        \includegraphics[trim=0 0 0 0, clip, width=1\textwidth]{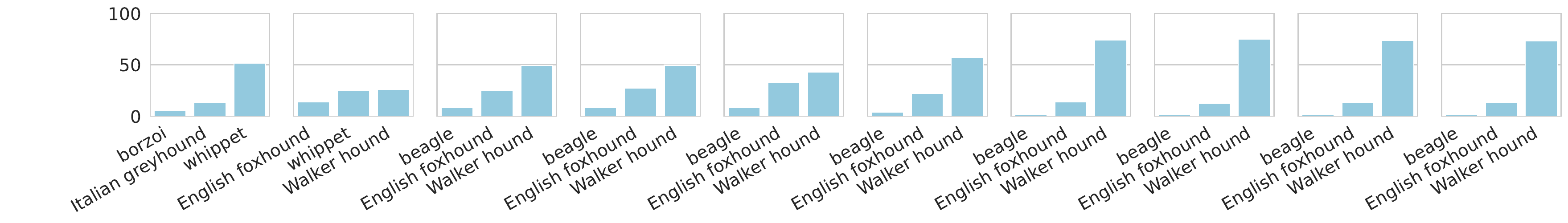}
        \caption{}    
        \label{fig:map1}
    \end{subfigure}
    \begin{subfigure}[b]{1\textwidth}
        \centering
        \includegraphics[trim=0 0 0 0, clip, width=1\textwidth]{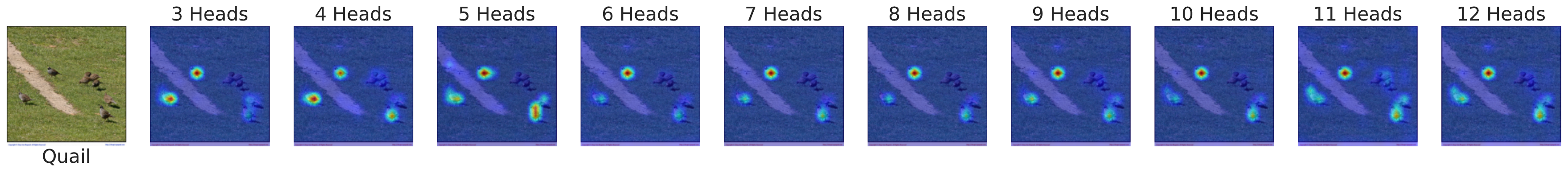}
        \includegraphics[trim=0 0 0 0, clip, width=1\textwidth]{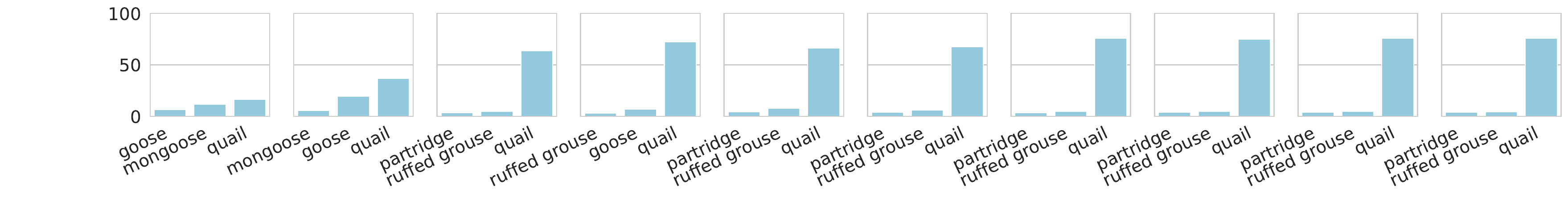}
        \caption{}    
        \label{fig:map2}
    \end{subfigure}
    \begin{subfigure}[b]{1\textwidth}
        \centering
        \includegraphics[trim=0 0 0 0, clip, width=1\textwidth]{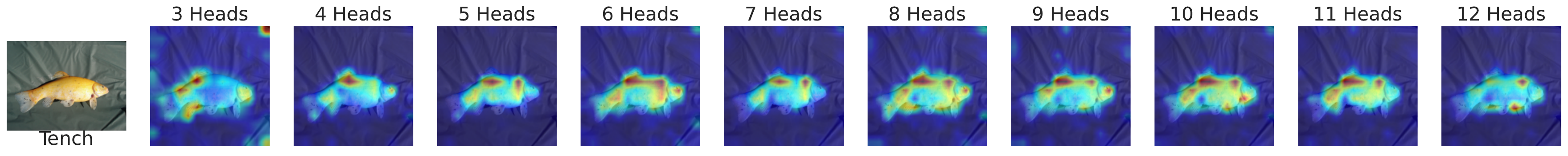}
        \includegraphics[trim=0 0 0 0, clip, width=1\textwidth]{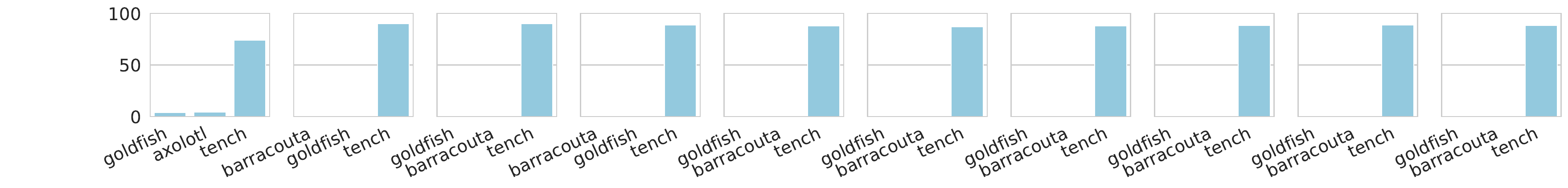}
        \caption{}    
        \label{fig:map3}
    \end{subfigure}
   \caption{Attention relevance maps~\citep{chefer2021transformer} of 3 samples from ImageNet-1K for \name with different number of heads. Increasing the number of heads leads to more confident classification and a more condensed attention distribution.}
   \label{fig:attention_map}
\end{figure*}

Fig.~\ref{fig:attention_map} displays the attention relevance map~\citep{chefer2021transformer} of selected subnetworks of \name, allowing us to visually investigate how the model's attention shifts when increasing the number of heads.
Fig.~\ref{fig:map3} shows that fewer heads lead to more scattered attention, whereas increasing the number of heads makes the attention maps more compact and focused on the main object.
Additionally, adding more heads enhances classification confidence. For instance, in Fig.~\ref{fig:map1}, the model misclassifies the input with 3 heads, but as we add more heads, the classification gradually shifts to the correct label and increases in confidence.
We also illustrate the t-SNE visualization of the final layer for different subnetworks, see Fig.~\ref{fig:tsne}. The figure shows that subnetworks with more heads exhibit a denser representation, while having fewer heads results in a more sparse representation. This indicates that increasing the number of heads enhances focus on the main object, which results in less entropy and, thereby, a more compact t-SNE representation. It is worth noting that the outliers in this figure occur due to the high norm values of the embeddings~\citep{darcet2024vision}.

\review{
\subsection{Robustness of \name}
To show the robustness of \name we also evaluate on different ImageNet variants: ImageNet-v2~\citep{recht2019imagenet}, ImageNet-R~\citep{hendrycks2020many}, ImageNet-A~\citep{hendrycks2019nae}, ImageNet-Sketch~\citep{wang2019learning}, and ImageNet-ReaL~\citep{beyer2020imagenet,ILSVRC15}. On four of these five ImageNet variants, \name achieves the overall best results. Figure~\ref{fig:GMAC_acc_imagenetv2} shows this for ImageNet-v2. Figure~\ref{fig:GMAC_acc_imagenetr} shows the only ImageNet variant, i.e., ImageNet-R, where \name trained with 9 to 12 heads is not able to achieve the best results. Nevertheless, \name reaches a competitive accuracy on these difficult variants and has on average the best results. See Appendix~\ref{sec:appendix_variants} for full results.

\begin{figure}
\begin{minipage}{.45\textwidth}
\centering
    \includegraphics[width=0.9\textwidth]{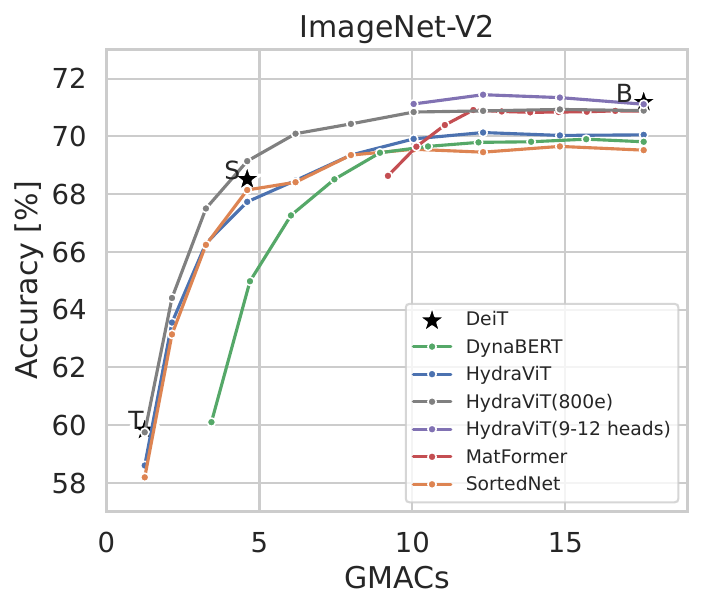}
    \caption{\review{Performance comparison of \name and baselines in terms of GMACs on ImageNet-v2.} \label{fig:GMAC_acc_imagenetv2}}
\end{minipage}
\hfill
\begin{minipage}{.45\textwidth}
\centering
    \includegraphics[width=0.9\textwidth]{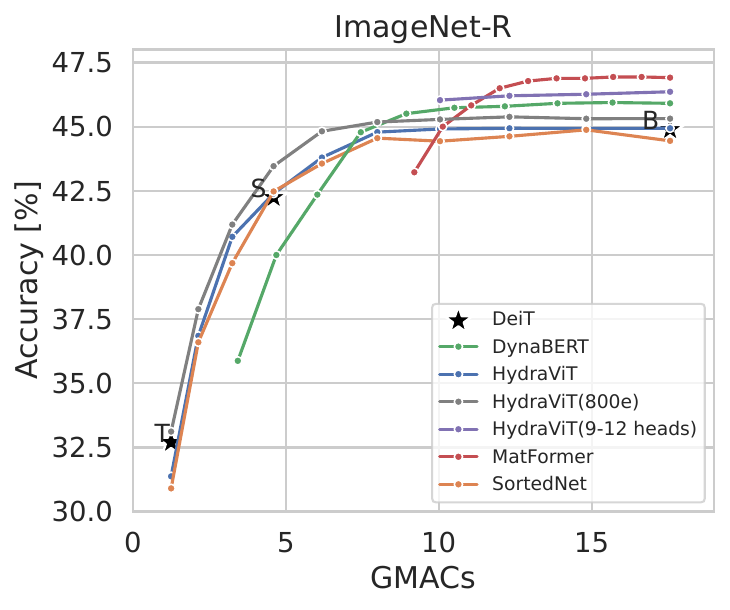}
\vspace{.4em}
    \caption{\review{Performance comparison of \name and baselines in terms of GMACs on ImageNet-R.} \label{fig:GMAC_acc_imagenetr}}
\end{minipage}
\end{figure}

\subsection{Limitations}
\label{sec:limitations}

\textbf{Training complexity} \name optimizes 10 loss functions simultaneously, which increases the computational load on the optimization progress. As a result, we require more training iterations to achieve accuracy comparable to that of individually trained models such as DeiT-tiny, DeiT-small, and DeiT-base. However, by training multiple models within a unified framework, \name ultimately requires much less total training time compared to training each of these 10 models for 300 epochs individually. See Appendix~\ref{sec:appendix_complexity} for more details on training complexity.

\textbf{Evaluation on different hardware} Our main focus with \name is on the efficiency and scalability on a single device, rather than the deployment on smaller hardware. However, metrics such as GMACs and params, are consistent across different platforms. Additionally, the skeleton of \name is identical to DeiT, and others have evaluated the latency and performance metrics of DeiT on smaller devices. For instance, FastViT~\citep{vasu2023fastvit} evaluates DeiT on the iPhone 12 Pro, MobileViT~\citep{mehta2023separable} on the iPhone 12, SPViT~\citep{kong2022spvit} on the ZCU102 FPGA and Galaxy S20, and GhostNetV3~\citep{liu2024ghostnetv3exploringtrainingstrategies} on the Huawei Mate 40 Pro. These studies provide insight into the expected performance and latency of \name on different hardware, indirectly supporting our claims about \name's adaptability.

\textbf{Evaluation on other models} While \name has been evaluated on DeiT-tiny, DeiT-small, and DeiT-base configurations, which have the same number of layers, we have not yet applied it to larger models like DeiT-large with more layers. We plan to explore this in future works.

}
\section{Conclusion}
\label{sec:conclusion}
We introduce \name, a novel approach for achieving a scalable \gls{ViT} architecture. By dynamically stacking attention heads and adjusting embedded dimensions within the \gls{MHA} layer during training, \name induces multiple subnetworks within a single model. This enables \name to adapt to diverse hardware environments with varying resource constraints while maintaining strong performance. Our experiments on ImageNet-1K demonstrate that \name achieves significant accuracy improvements compared to baseline approaches, with up to 5 percentage points higher accuracy at the same computational cost and up to 7 percentage points higher accuracy at the same throughput. This makes \name a practical solution for real-world deployments where hardware availability is diverse or changes over time.

\begin{ack}
\review{This research has received funding from the Federal Ministry for Digital and Transport under the CAPTN-F\"{o}rde 5G project grant no.~45FGU139\_H and the Federal Ministry for Economic Affairs and Climate Action under the Marispace-X project grant no.~68GX21002E. It was supported in part through high-performance computing resources available at the Kiel University Computing Centre.}
\end{ack}

\bibliographystyle{ACM-Reference-Format}
\bibliography{BibTex}

\appendix
\newpage
\section{Detailed Results of Submodels}
\label{sec:appendix_baselines}

\begin{table}[h!]
  \caption{Detailed results of all subnetworks for each Baseline and \name. Note that DynaBERT relies on knowledge distillation in every block, which is why it reaches less than 1\% accuracy when trained from scratch.}
  \label{tab:baselines_appendix}
  \centering
  \small
  \begin{tabular}{lrrrrrrr}
    \toprule
    \multirow{2}{*}{\textbf{Method}}&\multirow{2}{*}{\textbf{Dim}}&\textbf{RAM}&\textbf{MACs}&\textbf{Params}&\textbf{Throughput}&\textbf{Acc [\%]}&\textbf{Acc [\%]}\\
    &&\textbf{[MB]}&\textbf{[G]}&\textbf{[M]}&\textbf{[\# / s]}&\textbf{from scratch}&\textbf{from DeiT-tiny}\\
    \midrule
    \multirow{10}{*}{Matformer} & 768 & 508.45 & 17.56 & 86.6 & 1728.3 & 81.89 & 82.04 \\ 
                                & 704 & 484.73 & 16.63 & 81.8 & 1822.5 & 81.89  & 82.04 \\ 
                                & 640 & 461.0  & 15.7  & 77.1 & 1860.3 & 81.89 & 82.04 \\ 
                                & 576 & 437.27 & 14.78 & 72.4 & 1960.1 & 81.87 & 81.99 \\ 
                                & 512 & 413.55 & 13.85 & 67.7 & 2020.7 & 81.79 & 81.95 \\ 
                                & 448 & 389.81 & 12.92 & 63.0 & 2128.4 & 81.66 & 81.94 \\ 
                                & 384 & 366.08 & 11.99 & 58.2 & 2231.6 & 81.52  & 81.80 \\ 
                                & 320 & 342.36 & 11.06 & 53.5 & 2356.4 & 81.09  & 81.71 \\ 
                                & 256 & 318.63 & 10.13 & 48.8 & 2444.8 & 80.47 & 81.25 \\  
                                & 192 & 294.9  & 9.2   & 44.1 & 2601.4 & 79.40 & 80.48  \\
    \midrule
    \multirow{10}{*}{DynaBERT}  & 768 & 508.45 & 17.56 & 86.6 & 1725.7 & - & 81.30   \\ 
                                & 704 & 471.65 & 15.68 & 79.5 & 1876.8 & - & 81.29 \\ 
                                & 640 & 434.85 & 13.88 & 72.4 & 1995.7 & - & 81.20 \\ 
                                & 576 & 398.03 & 12.16 & 65.3 & 2213.7 & - & 81.22 \\ 
                                & 512 & 361.23 & 10.51 & 58.2 & 2412.4 & - & 81.05 \\ 
                                & 448 & 324.43 & 8.94 & 51.2 & 2709.3 & - & 80.71 \\ 
                                & 384 & 287.62 & 7.45 & 44.0 & 3014.6 & - & 80.16 \\ 
                                & 320 & 250.81 & 6.03 & 37.0 & 3522.5 & - & 78.78 \\  
                                & 256 & 214.01 & 4.69 & 30.0 & 4157.1 & - & 77.04 \\  
                                & 192 & 177.2  & 3.43  & 22.8 & 4944.5 & - & 73.00 \\
    \midrule
    \multirow{10}{*}{SortedNet} & 768 & 508.45 & 17.56 & 86.6 & 1753.0 & 79.71 & 80.80 \\ 
                                & 704 & 440.19 & 14.82 & 72.7 & 1629.5 & 79.79 & 80.70 \\ 
                                & 640 & 376.63 & 12.31 & 60.3 & 1846.6 & 79.82 & 80.74 \\ 
                                & 576 & 317.8  & 10.04 & 49.0 & 2318.3 & 79.69 & 80.60 \\ 
                                & 512 & 263.68 & 7.99  & 38.8 & 2466.7 & 79.28 & 80.43 \\ 
                                & 448 & 214.27 & 6.18  & 29.9 & 2612.1 & 78.88 & 79.93 \\  
                                & 384 & 169.6  & 4.6   & 22.1 & 3874.8 & 77.79 & 78.94 \\ 
                                & 320 & 129.63 & 3.25  & 15.4 & 3886.0 & 75.85 & 77.55 \\ 
                                & 256 & 94.4   & 2.14  & 10.0 & 4654.6 & 72.55 & 74.92 \\ 
                                & 192 & 63.87  & 1.25  & 5.7  & 5898.2 & 66.64 & 70.20   \\
    \midrule
    \multirow{10}{*}{\name}     & 768 & 508.45 & 17.56 & 86.6 & 1754.1 & 80.45 & 81.10 \\ 
                                & 704 & 440.19 & 14.82 & 72.7 & 1916.1 & 80.47 & 81.08 \\ 
                                & 640 & 376.63 & 12.31 & 60.3 & 2242.9 & 80.42 & 81.03 \\ 
                                & 576 & 317.8  & 10.04 & 49.0 & 2503.2 & 80.36 & 80.99 \\
                                & 512 & 263.68 & 7.99  & 38.8 & 3141.5 & 80.03 & 80.80 \\ 
                                & 448 & 214.27 & 6.18  & 29.9 & 3616.4 & 79.45 & 80.35 \\ 
                                & 384 & 169.6  & 4.6   & 22.1 & 4603.6 & 78.40 & 79.28 \\
                                & 320 & 129.63 & 3.25  & 15.4 & 5652.0 & 76.66 & 77.80 \\ 
                                & 256 & 94.4   & 2.14  & 10.0 & 7558.2 & 73.39 & 75.40 \\ 
                                & 192 & 63.87  & 1.25  & 5.7  & 10047.6 & 67.34 & 70.56  \\
    \midrule
                                & 768 & 508.45 & 17.56 & 86.6 & 1754.1 & \review{81.93} & 81.60 \\ 
                                & 704 & 440.19 & 14.82 & 72.7 & 1916.1 & \review{81.90} & 81.57 \\ 
    \multirow{5}{*}{\name}      & 640 & 376.63 & 12.31 & 60.3 & 2242.9 & \review{81.84} & 81.63 \\ 
    \multirow{5}{*}{800 Epochs} & 576 & 317.8  & 10.04 & 49.0 & 2503.2 & \review{81.73} & 81.48 \\
                                & 512 & 263.68 & 7.99  & 38.8 & 3141.5 & \review{81.54} & 81.36 \\ 
                                & 448 & 214.27 & 6.18  & 29.9 & 3616.4 & \review{80.98} & 80.92 \\ 
                                & 384 & 169.6  & 4.6   & 22.1 & 4603.6 & \review{79.84} & 80.15 \\
                                & 320 & 129.63 & 3.25  & 15.4 & 5652.0 & \review{78.07} & 78.43 \\ 
                                & 256 & 94.4   & 2.14  & 10.0 & 7558.2 & \review{74.83} & 75.95 \\ 
                                & 192 & 63.87  & 1.25  & 5.7  & 10047.6 & \review{68.78} & 71.67 \\
    \bottomrule
  \end{tabular}
\end{table}

\review{

\section{Generalization of Submodels}
Figure~\ref{fig:trainingcurves} shows that we do not overfit on any subnetwork even when training for 800 epochs. Instead, in \name, stochastic dropout training minimizes the loss more or less uniformly across all subnetworks. This is especially true when we have more subnetworks, as only one subnetwork is optimized per batch and our loss objective gets more complicated. Therefore \name needs more epochs to reach a specific accuracy with a subnetwork than the individually trained subnetwork.

\begin{figure}[t]
    \centering
    \includegraphics[width=\linewidth]{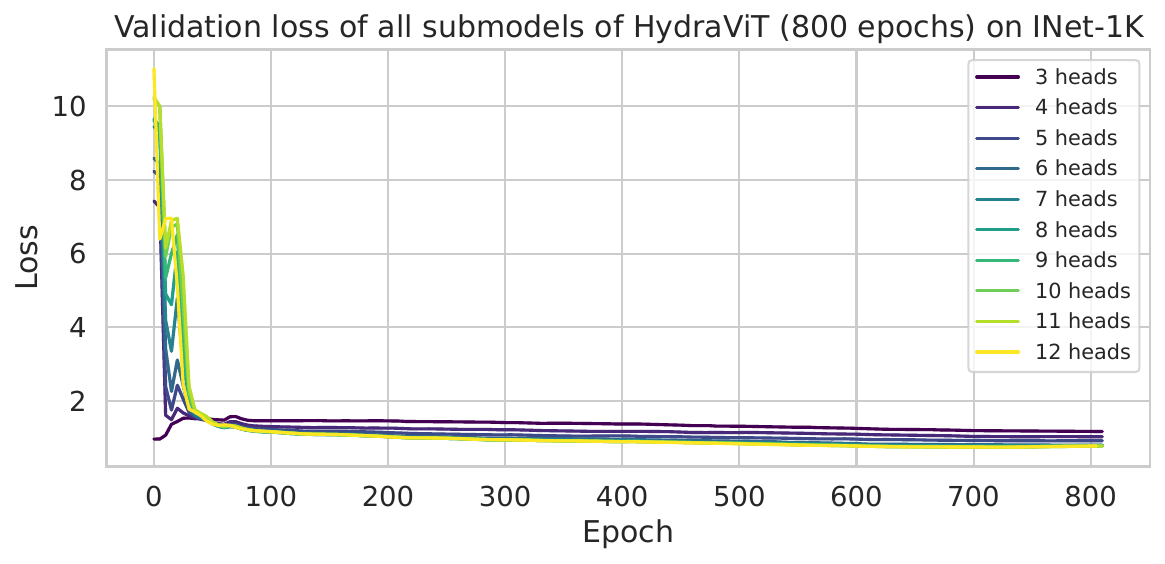}
    \caption{\review{Training loss for all subnetworks when training \name for 800 epochs. \label{fig:trainingcurves}}}
\end{figure}

\section{Training Complexity}
\label{sec:appendix_complexity}
Table~\ref{tab:training-complexity} shows that we need to take model size into account when comparing training epochs. While DeiT-tiny is by far the fastest to complete one epoch, \name is actually on average faster to train with its 10 subnetworks than DeiT-small and DeiT-base.

\begin{table}[t]
    \caption{Average training throughput of \name and DeiT models for one epoch. }
    \centering
    \vspace{0.4em}
    \begin{tabular}{cr}
        \Xhline{2\arrayrulewidth} \\[-2ex]
        \textbf{Model} & \textbf{Throughput [\#/s]} \\ 
        \Xhline{2\arrayrulewidth}\\[-2ex]
        
        DeiT-tiny  & 1610 \\
        DeiT-small & 821  \\
        DeiT-base  & 340  \\
        \name      & 1093 \\

        \Xhline{2\arrayrulewidth}  & \\[-2ex]
    \end{tabular}
    \label{tab:training-complexity}
\end{table}

\section{Model Loading Time}
Table~\ref{tab:loadingtime} shows that the loading time, i.e., the time it takes to load a model into RAM, is for many applications low enough to enable model switching during runtime.

\begin{table}
    \centering
    \caption{HydraViT loading times, each model was loaded six times.}
    \begin{tabular}{cccc}
        \Xhline{2\arrayrulewidth} \\[-2ex]
        \textbf{Model} & \textbf{Heads} & \textbf{Dim} & \textbf{Latency [ms]} \\ 
        \Xhline{2\arrayrulewidth}\\[-2ex]
        HydraViT & 3  & 192 & 74.1  $\pm$ 1.4 \\
        HydraViT & 6  & 384 & 133.3 $\pm$ 6.1\\  
        HydraViT & 12 & 768 & 138.6 $\pm$ 5.8\\
        \Xhline{2\arrayrulewidth}\\[-2ex]
    \end{tabular}
    \label{tab:loadingtime}
\end{table}

\section{Initializing \name with DeiT-base instead of DeiT-tiny}
In \name, the attention heads are treated as stacks, with each head built on top of the previous ones. The smallest submodel, with 3 heads, is equivalent to DeiT-tiny. Initializing with DeiT-tiny positions this 3-head submodel at its optimal starting point and ensures that larger submodels, which also contain these 3 heads, also start from a good local optimum. By employing the stochastical dropout training introduced in \name, additional heads (4th, 5th, etc.) are trained iteratively on top of the initial three heads, creating a layered stack of attention heads. DeiT-base initialization helps \name at 12 heads but does not yield strong smaller submodels, as Table~\ref{tab:deitb} shows. 

\begin{table}
    \caption{The accuracy of \name when initialized with DeiT-tiny vs DeiT-base. While the accuracy at the 12 heads is higher with DeiT-base initialization the average accuracy is lower than with DeiT-tiny initialization.}
    \centering
    \vspace{0.4em}
    \begin{tabular}{ccrrr}
        \Xhline{2\arrayrulewidth} \\[-2ex]
        \multirow{2}{*}{\textbf{Initialization}} & \multirow{2}{*}{\textbf{Epochs}} & \textbf{Acc [\%]} & \textbf{Acc [\%]} & \textbf{Acc [\%]} \\ 
        & & \textbf{3 Heads} & \textbf{6 Heads} & \textbf{12 Heads} \\ 
        \Xhline{2\arrayrulewidth}\\[-2ex]
        
        DeiT-tiny & 400 & 73.16 & 79.63 & 80.90 \\
        DeiT-base & 400 & 70.83 & 79.62 & 81.84 \\

        \Xhline{2\arrayrulewidth}  & \\[-2ex]
    \end{tabular}
    \label{tab:deitb}
\end{table}

\section{Results on ImageNet variants}
\label{sec:appendix_variants}
Figure~\ref{fig:full-results} shows all results of \name and baselines in terms of GMACs and throughput on all ImageNet variants. Note that Figure~\ref{fig:GMAC_acc_imagenetv2} and Figure~\ref{fig:GMAC_acc_imagenetr} contain the results for accuracy per GMACs on ImageNet-v2 and ImageNet-R.

\begin{figure}
    \centering
    
    \begin{minipage}{0.44\textwidth}
        \centering
        \includegraphics[width=\textwidth]{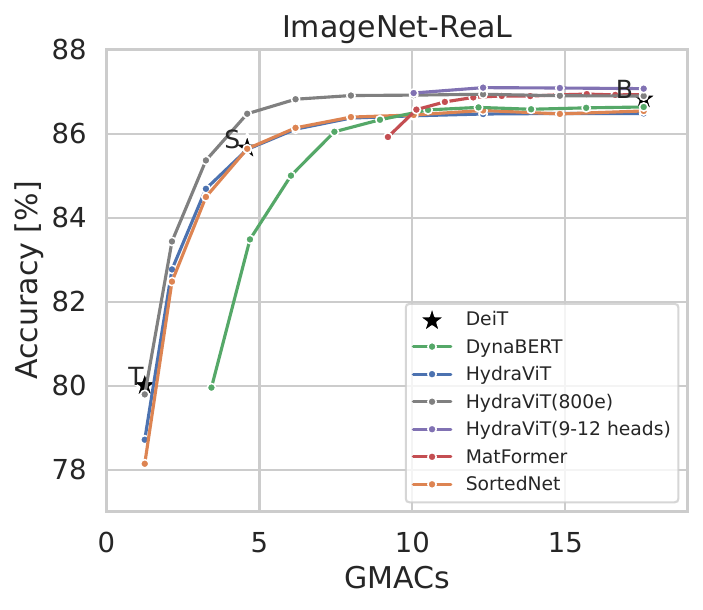}
    \end{minipage}\hfill
    \begin{minipage}{0.44\textwidth}
        \centering
        \includegraphics[width=\textwidth]{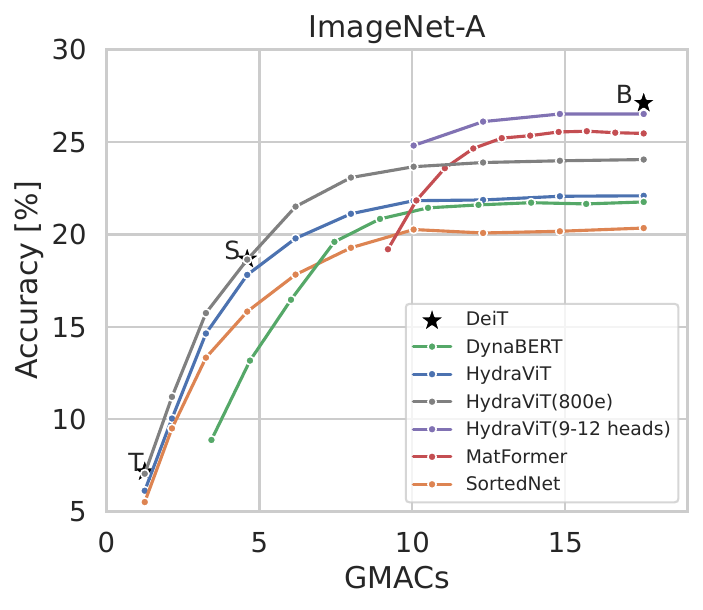}
    \end{minipage}

    \begin{minipage}{0.44\textwidth}
        \centering
        \includegraphics[width=\textwidth]{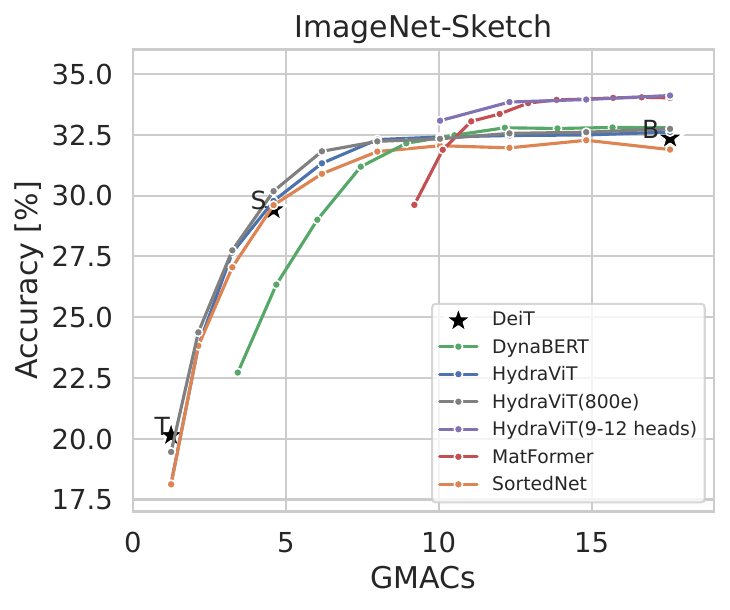}
    \end{minipage}\hfill
    \begin{minipage}{0.44\textwidth}
        \centering
        \includegraphics[width=\textwidth]{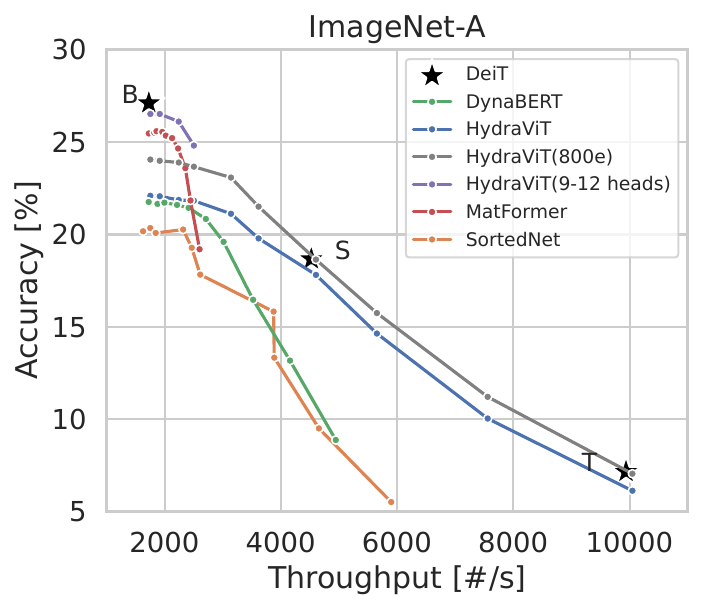}
    \end{minipage}

    \begin{minipage}{0.44\textwidth}
        \centering
        \includegraphics[width=\textwidth]{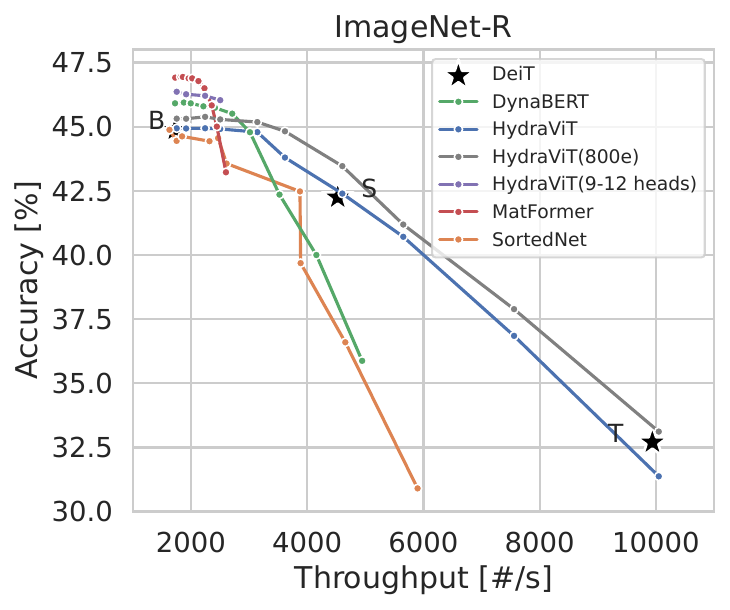}
    \end{minipage}\hfill
    \begin{minipage}{0.44\textwidth}
        \centering
        \includegraphics[width=\textwidth]{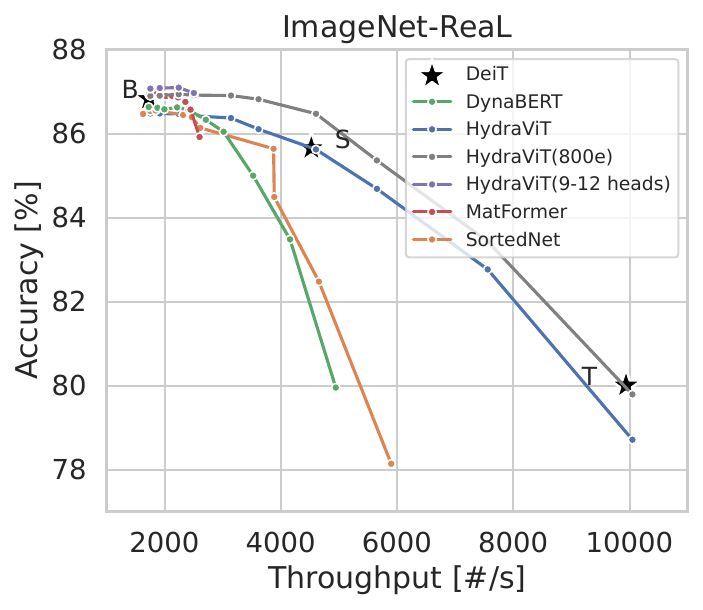}
    \end{minipage}

    \begin{minipage}{0.44\textwidth}
        \centering
        \includegraphics[width=\textwidth]{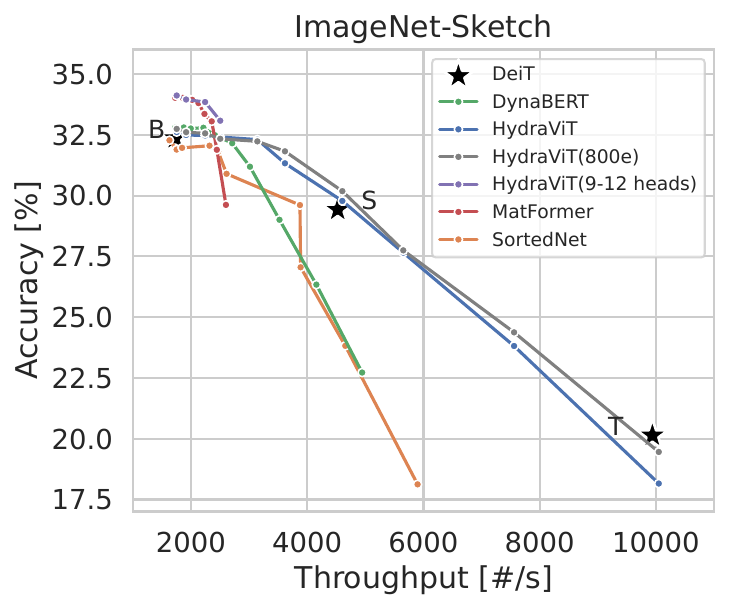}
    \end{minipage}\hfill
    \begin{minipage}{0.44\textwidth}
        \centering
        \includegraphics[width=\textwidth]{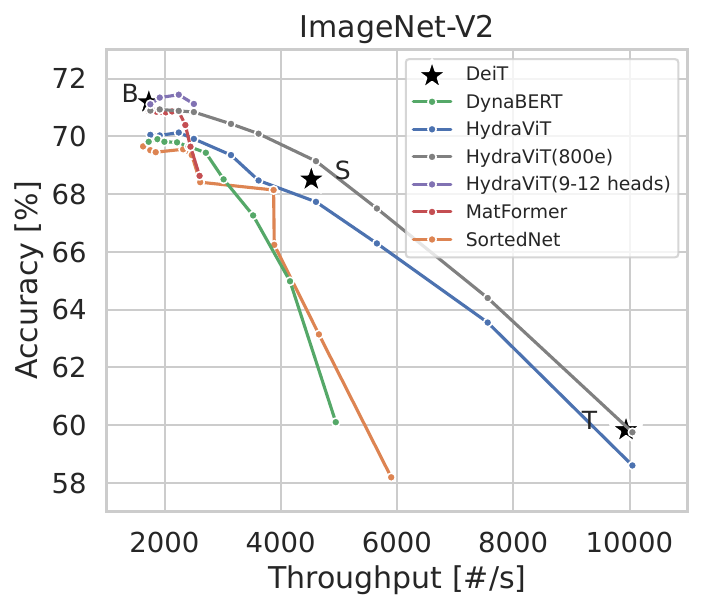}
    \end{minipage}
    
    \caption{\review{Full results of \name and baselines in terms of GMACs and throughput on ImageNet variants.} \label{fig:full-results}}
\end{figure}

}

\end{document}